\documentclass[journal]{IEEEtran}

\ifCLASSOPTIONcompsoc
\usepackage[nocompress]{cite}
\else
\usepackage{cite}
\fi

\usepackage{booktabs}
\usepackage{multirow}

\ifCLASSINFOpdf
\usepackage[pdftex]{graphicx}
\graphicspath{{Block_Diagrams/}}
\else

\fi

\usepackage{amsmath}

\usepackage{array}

\newcolumntype{P}[1]{>{\centering\arraybackslash}p{#1}}
\newcolumntype{M}[1]{>{\centering\arraybackslash}m{#1}}

\usepackage{stfloats}

\usepackage[justification=centering]{caption}
\usepackage{tabularx}

\usepackage[hidelinks]{hyperref}

\begin{document}
	%
	\title{SwishNet: A Fast Convolutional Neural Network for Speech, Music and Noise\\Classification and Segmentation}
	%
	%
	%
	%
	
	\author{Md.~Shamim~Hussain,
		and~Mohammad~Ariful~Haque
		\thanks{ M. S. Hussain and M. A. Haque were with the Department
			of Electrical and Electronic Engineering, Bangladesh University of Engineering and Technology, Dhaka, Bangladesh.\protect\\
			E-mail: snirjhar@gmail.com, arifulhoque@eee.buet.ac.bd}
	}

	\IEEEtitleabstractindextext{%
		\begin{abstract}
			Speech, Music and Noise classification/segmentation is an important preprocessing step for audio processing/indexing. To this end, we propose a novel 1D Convolutional Neural Network (CNN) - SwishNet. It is a fast and lightweight architecture that operates on MFCC features which is suitable to be added to the front-end of an audio processing pipeline. We showed that the performance of our network can be improved by distilling knowledge from a 2D CNN, pretrained on ImageNet. We investigated the performance of our network on the MUSAN corpus - an openly available comprehensive collection of noise, music and speech samples, suitable for deep learning. The proposed network achieved high overall accuracy in clip (length of 0.5-2s) classification (\textgreater97\% accuracy) and frame-wise segmentation (\textgreater93\% accuracy) tasks with even higher accuracy (\textgreater99\%) in speech/non-speech discrimination task. To verify the robustness of our model, we trained it on MUSAN and evaluated it on a different corpus –- GTZAN and found good accuracy with very little fine-tuning. We also demonstrated that our model is fast on both CPU and GPU, consumes a low amount of memory and is suitable for implementation in embedded systems.
		\end{abstract}
		
		}

	\maketitle

	\IEEEdisplaynontitleabstractindextext

	%
	\IEEEpeerreviewmaketitle

	\section{Introduction}\label{sec:introduction}

	%
	%
	%
	%
	\IEEEPARstart{A}{udio} classification involves assigning the content of a given audio excerpt to a particular class and audio segmentation involves assigning different temporal regions of a media to different classes. Music, speech and noise classification and segmentation is an important task because these three types of signals are inherently different in nature \cite{Wolfe2002} and require different types of processing and/or coding schemes. For example, a good compression scheme for speech may not be good for compressing music. Also, the signal processing steps used for these two types of media are likely to be very different. So, correctly identifying the type of media before further processing should be considered a very important task.

	In speech processing and coding, we are interested in distinguishing speech and non-speech regions in a given signal, which is the objective of Voice Activity Detection (VAD) algorithms \cite{Kola2011}. Although many conventional VAD algorithms perform well in distinguishing voice from the background, they are prone to making errors when the same media contains vocal music and speech, such as TV and radio broadcast. Moreover, some types of noises (e.g. babble noise) may seem to have speech like features when looked at over only a short period of time e.g. a single frame or only a few consecutive frames. Therefore, it is desirable to formulate a method that efficiently uses contextual information i.e. the information from previous frames, to distinguish speech from both non-speech signals i.e. music and background noise.

	On the advent of widespread use of the internet, it is now possible to build large media databases from user-contributed data. However, labels for the collected data are rarely available. Manual media labeling can be both expensive and time-consuming. So, a reliable method is required to automate the indexing of large media databases. It is intuitive to first classify/segment the media into broad categories such as speech, music and noise which may be followed by further fine-grained classification/segmentation procedure (for example, into different musical genres).
	
	\begin{figure*}[!t]
		\centering
		\includegraphics[width=\textwidth]{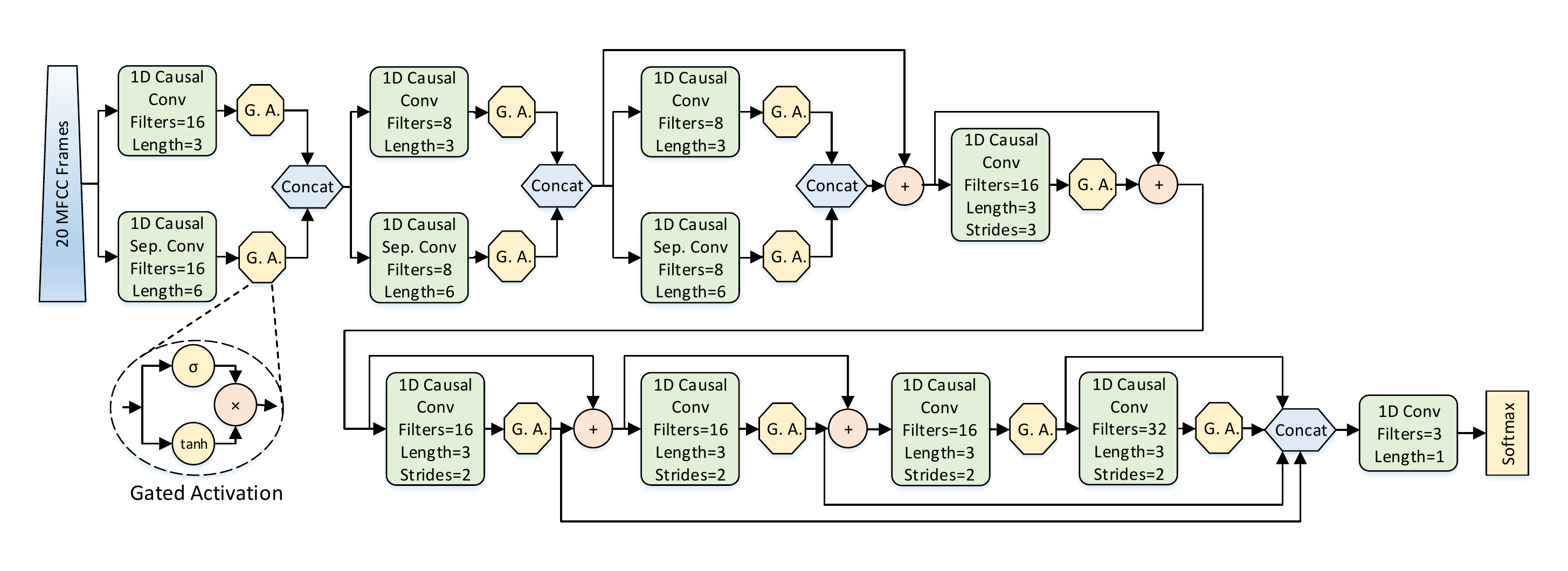}
		\caption{Network Architecture of SwishNet-slim}
		\label{fig:swishnet}
	\end{figure*}

	The above discussion suggests that it is desirable to bring the classification/segmentation of media into speech, music and noise under a single framework. Engineering appropriate features for this problem can be both difficult and cumbersome. For example, a capella music has features similar to speech, whereas rock and roll music may have features similar to noise. Moreover, musical trends change over time and new genres may appear. So, we tried to solve this problem from general purpose features (such as MFCC and spectrogram) rather than artificially formulated task-specific features. Also, we wanted to use the contextual information most effectively. In recent years deep learning \cite{LeCun2015} algorithms have achieved unprecedented success in numerous classification problems, without any need for careful feature selection. Especially, deep convolutional neural networks have set new frontiers in many fields such as image and audio classification and segmentation. However, deep neural networks are generally known to be more computationally expensive and slower than other more conventional models. These models often need large-scale parallelization in GPUs for faster implementation. Since classification/segmentation is often set as a pre-processing step in an audio processing pipeline, it needs to be fast enough to prevent the delay in further processing. Also, the computational resources are limited in many scenarios such as in mobile devices and embedded systems where it is unreasonable to waste too much resource in the preprocessing step. This is where we put forth our contribution. In this work, we present SwishNet, a carefully designed novel 1D convolutional deep neural network architecture which can achieve a high level of accuracy while also being fast, lightweight and memory efficient, even without large-scale parallelization in a GPU. Also, the deep convolutional nature of the architecture allows it to pick up contextual information from previous frames effectively. It can be trained on a sufficiently diverse dataset such as the MUSAN corpus without the need for any specific feature engineering. And in the presence of new data (e.g. new musical genres), it can simply be fine-tuned without the need for retraining from scratch.
	
	
	
	\section{Proposed Network Architecture}\label{sec:proparch}
	\subsection{A Review of Deep Convolutional Neural Networks}
	A Deep Neural Network (DNN) \cite{Bengio2009} consists of multiple layers of neurons stacked on top of one another. Higher layers process the outputs from the lower layers and thus form a more abstract representation of the input data. Multiple layers of abstraction allow the network to extract complex features inherently. Convolutional Neural Networks (CNN) \cite{LeCun1995, Krizhevsky2012} are a class of feed-forward neural networks, which implement one or more convolutional layers. A convolutional layer consists of a set of filters, each of which shares its weights across spatial/temporal dimensions of the input. This ensures a shift-invariant approach which is effective for feature extraction. CNNs also have fewer parameters than fully connected networks because of weight sharing.
	
	\subsection{Feature Selection}
	While 1D convolutional networks, like WaveNet \cite{VanDenOord2016b} can operate on raw audio, we trained our network on frame-wise extracted features because processing raw audio input is infeasible in terms of computational cost and memory requirement for our problem. 
	Since the discrimination of speech, noise and music is related to human perception, we investigated perceptually weighted features. MFCC (Mel-Frequency Cepstral Coefficients) is the most versatile feature which is used in both speech and music \cite{Logan2000} processing. So, we used frame-wise MFCC for training our network.
	
	\subsection{SwishNet – A 1D Convolutional Neural Network}
	SwishNet is a 1D convolutional network which operates on frame-wise MFCC features. The rationale behind designing a 1D CNN instead of a conventional 2D CNN (that operates on spectrogram) is that, 1D convolutions are much less computationally expensive and 1D feature maps require less memory during processing. Also, a carefully designed 1D CNN is likely to be much smaller in size than a 2D CNN. In SwishNet, convolutions are carried out along the temporal dimension only. The MFC coefficients are treated as input channels. This approach is based on the fact that MFC coefficients are are almost uncorrelated, like the channels (i.e. Red, Green, Blue) of an image.

	We investigated with two versions of SwishNet i.e. slim and wide. The detailed network architecture of SwishNet-slim is shown in Fig. \ref{fig:swishnet}. For SwishNet-wide, we simply doubled the width of each layer.

	Our network architecture is partly inspired by the new WaveNet \cite{VanDenOord2016b} architecture and the Inception \cite{Szegedy2015} architecture. Like the WaveNet architecture, we used multiple layers of causal convolutions to gradually increase the receptive field and gated activation functions \cite{VandenOord2016a} containing sigmoid and tanh functions instead of widely used ReLU activations. The gated activations allow the network to select which information to pass from one time step to the next, just like a gated recurrent network. It also conveniently cuts down the number of feature maps passing from one layer to the next to half.

	The WaveNet architecture was originally designed for autoregressive audio generation, but our network is focused on classification. So, we used strided convolutions instead of dilated convolutions to reduce the computational cost. Adding residual and skip connections improved both accuracy and ease of training. Residual connections were applied to the middle layers and skip connections were applied to the last three layers.

	The WaveNet architecture was designed for audio signals with high sampling rates, so an exponential increase of the receptive field was encouraged. But, our network operates on derived features, so we used longer filters with overlapping receptive fields and allowed a more gentle increase of the receptive field from lower to upper layers in order to prioritize the capture of more recent information in the lower layers.

	We found that it is vital to capture long-term dependencies i.e. information in the previous frames in the lower layers, which requires longer filters. The inception architecture uses parallel branches to capture features at different scales. With a similar purpose in mind, we added depthwise separable 1D convolutional \cite{Chollet2016} layers with longer (kernel size ~6) un-dilated filters parallel to the conventional convolutional filters and concatenated the outputs from the two branches. Separable convolutions are faster and require less memory than conventional convolutions, especially for longer filter kernels, so this modification further improved accuracy without compromising size and speed.

	As a regularization method, we (optionally) added dropout layers (not shown in the figure) in between convolutional blocks which slightly improved accuracy. However, depending on the size of the dataset and the width of the network, it may not be necessary.
	
	\subsection{A 2D Convolutional Neural Network for Distillation and Comparison - MobileNet}
	The most widely used deep learning technique for audio classification is to apply a 2D convolutional network on the spectrogram or other derived features \cite{Sercu2016,Hershey2017,Takahashi2016}. However, to the best of our knowledge, no other deep learning scheme has specifically addressed classification/segmentation into all of the three classes - Music, Speech and noise under a single unified framework. So, for comparison, we sought an established 2D architecture that achieves highest performance for the task at hand. We propose that our 1D CNN can come very close in terms of performance to 2D CNNs while also drastically reducing the computational time, network size and memory requirement. We will also show that the knowledge from a 2D network can be distilled in SwishNet to further improve its performance.

	For 2D convolutional networks, we treated the log MFB (Mel Filter Banks) spectrum as a 2D signal and convolutions were done along both frequency and time axes. Among the CNN architectures, we considered, the ones that performed well on ImageNet \cite{Deng2009}, also performed well for our problem, especially when they were initialized with ImageNet weights. They also converged much faster and achieved higher accuracy when initialized with ImageNet weights rather than with random weights. So, it is apparent that transfer learning \cite{Oquab2014} is at work here, even though the input is not a natural image, but rather a visual representation of the time-frequency spectrum.

	To apply the pre-trained networks on log MFB spectrum, we copied the input along three input channels (Red, Green and Blue) which is equivalent to using a greyscale image. We only kept the convolutional parts of the networks and applied Global Average Pooling to the outputs. Then, we applied two dense layers ending in a softmax output.

	We experimented with VGG16 \cite{Simonyan2014}, MobileNet \cite{Howard2017} and NASNet Mobile \cite{Zoph2017}. We did not experiment with bigger networks such as ResNet or Inception because they would be unnecessarily large for our problem. Among these networks, MobileNet with a width multiplier equal to 0.25 (α=0.25) performed best in terms of accuracy, network size and speed.
	\begin{figure}[!t]
		\centering
		\includegraphics[width=\columnwidth]{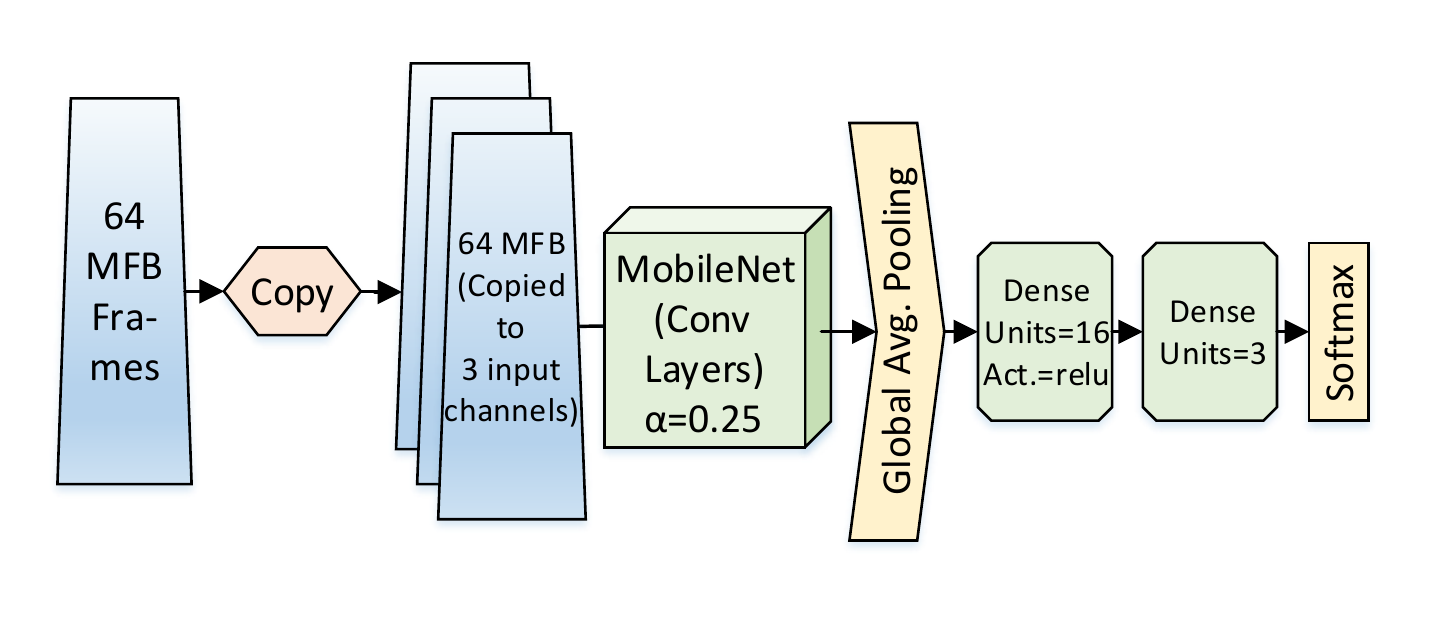}
		\caption{Network Architecture of MobileNet}
		\label{fig:mobilenet}
	\end{figure}

	MobileNet uses depthwise separable 2D convolutions which make the network fast while also keeping the network size in check. Details of the network can be found in \cite{Howard2017}. Fig. \ref{fig:mobilenet} shows the overview of our network architecture.

	As shown in \cite{Hinton2015}, knowledge from a bigger network can be distilled into a smaller network to improve its performance. Accordingly, we distilled the knowledge from frame-wise classification results produced by MobileNet to SwishNet which further improved the accuracy of SwishNet. We trained two versions of SwishNet with different widths. We found that a higher distillation temperature is suitable for the wider network which is consistent with \cite{Hinton2015}.
	
	\section{Experimental Methods}
	\subsection{Description of the Corpora}
	The MUSAN corpus \cite{Snyder2015} is compiled from Creative Commons and US Public Domain sources. It is freely available (at OpenSLR \cite{musanpage}) and redistributable. It consists of approximately 109 hours of audio in three categories. It has 660 music files of different genres, 930 noise files and 426 speech files (in 12 different languages) from LibriVox recordings and US Govt. hearings. All the audio files are in 16 kHz WAV format.

	The GTZAN \cite{Tzanetakis2002} music and speech collection contains 60 music files of different genres and 60 speech files from different sources. Each of the tracks is 30 seconds long. All files are in 22050Hz WAV format. 
	
	\subsection{Evaluation Strategy}
	The files in the MUSAN corpus were randomly divided into three sets. 65\%, 10\% and 25\% of the all files in each category (i.e. speech, music and noise) were assigned to training, validation and test sets respectively. The same preprocessing and data preparation steps were followed for all three sets. Silent parts of the files were removed using power thresholding. Loudness was equalized with a window length of 250ms. For speech/music classification on the GTZAN corpus, background noise was removed using the Log MMSE \cite{Ephraim1985} method. For evaluation on GTZAN, we selected 25\% of the files for fine-tuning and the rest for testing. For classification data, we segmented the files to form either 0.5s, 1s or 2s clips with 50\% overlap. The clips were framed into 25ms frames with 15ms overlap. 20 MFC coefficients from 32 mel-frequency bands or 64 band log MFB features were extracted from each frame depending on the model to be trained.

	For segmentation, we used the same networks trained for classification purpose. We built an artificial dataset of 500 files of randomly chosen lengths ranging 20s to 120s; by randomly concatenating silence, noise, speech and music signals from the test set with average lengths 0.5s, 5s, 10s and 12s respectively. Some of the transitions were abrupt, while others were separated by silence. Instead of zeroing out the silenced parts, we used natural silence portions clipped from within the dataset. We applied the neural networks on a windowed signal around a particular frame to gather contextual information. So, the decision of a frame depends not only on that frame but also on other frames within the contextual window. Window length was varied from 0.5s to 2.0s in order to evaluate its impact on the segmentation performance. Then, median filtering (filter length of 200 frames) was applied on the frame-wise predictions (probabilities) which slightly improved final accuracy. Accuracy was measured frame-wise against true annotations. Silence frames were excluded from the evaluation.

	As a baseline, we compared our results with the widely used Gaussian Mixture Model (GMM) and also with a Fully-Connected Neural Network (FNN). They were both trained in a frame-wise manner, and MFCC with first and second order deltas were used as input features. The final decision for segment classification was taken by majority voting. We used 3 GMMs for the 3 classes, each with 256 components with diagonal covariance. As for the FNN, we used a 4-layered Self-Normalizing Neural Network (SNN) \cite{Klambauer2017} with Alpha dropout (which is the current state-of-the-art for FNNs). The design choices of these models were focused on improving classification performance rather than speed.
	
	\subsection{Optimization Strategy}
	We used the Adam optimizer \cite{Kingma2014} to train all our networks. Gated activations used in SwishNet seemed to cause vanishing gradient problems to some extent and training was relatively slow. So, to speed up training we used cosine annealing and warm restarts as described in \cite{Loshchilov2016}. The batch size was tuned for fastest training. For smaller segments, the number of training samples increased, so we increased the batch size proportionately to keep the number of gradient updates per epoch same. We trained SwishNet for 120 epochs. MobileNet trained really fast when initialized with ImageNet weights and reached convergence within very few (8-10) epochs. The GMM was trained until convergence and SNN was trained until the validation loss plateaued. For fine-tuning on GTZAN we took the networks trained on MUSAN, and trained them with a low learning rate for 50 epochs on the GTZAN training data.
	
	\subsection{Distillation Strategy}
	For distillation a pretrained (on ImageNet) MobileNet was first trained on the training data. Then we used the output logits from the MobileNet to distill SwishNet (on the same training data). The validation data was used to choose the best temperature for distillation. We formed a weighted loss from 90\% of the soft target loss and 10\% of the true label loss. The best temperatures were found to be 4, 2, 1 for SwishNet-sim and 10, 10, 8 for SwishNet-wide for clip lengths of 2s, 1s and 0.5s respectively.
	
	\section{Results and Discussions}
	
	\subsection{Classification Results}
	\begin{figure*}[!t]
		\centering
		\includegraphics[trim={1em 1em 1em 1em},clip, width=\linewidth]{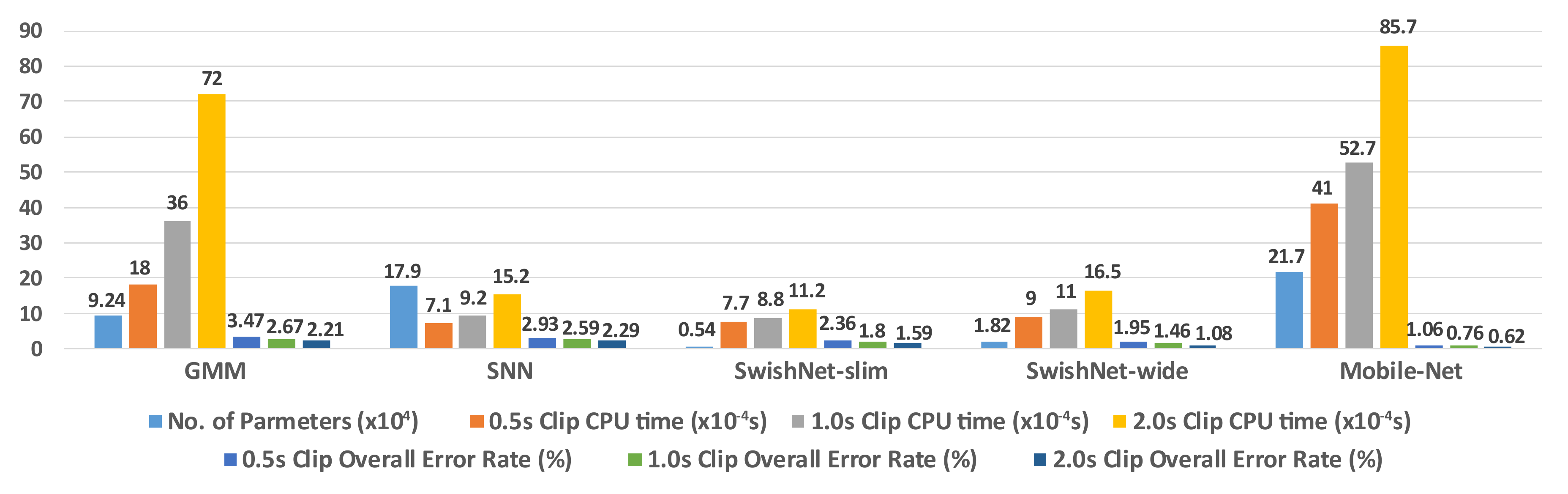}
		\caption{Size, speed and performance comparison of the models (lower is better for all parameters)}
		\label{fig:perfchrat}
	\end{figure*}
	Both SwishNet and MobileNet can handle variable input lengths. So we kept the network structures fixed for all tests. Table \ref{tab:speed} compares the sizes prediction speeds of the three networks as well as the baseline models. To simulate comparison of real-time computational cost, we fixed batch size to 1 for all models and compared their speed on CPU and GPU. Our CPU was Core i7 8700K, a 6 core processor and our GPU was NVIDIA GTX 1080ti and the networks were implemented in Tensorflow \cite{Abadi2016}. We see that SwishNet-slim is actually faster on CPU than on GPU and so is SwishNet-wide for shorter clip lengths. This is due to the fact that, when parallelization requirement is minimal, clock speed plays the major role in computational speed. Also, latency is introduced when data is passed from CPU memory to GPU memory. SwishNet slim has a computational speed on CPU comparable to SNN while being much smaller in size. The fully-connected SNN enjoys an advantage in speed performance due to its lower depth and simplistic computational requirement. But, as we will show presently, the performance gain of SwishNet justifies the use of a deep convolutional architecture. SwishNet is 4 to 8 times faster than MobileNet on CPU, due to its low parallelization requirement.


\begin{table}[!t]
	\centering
	\caption{Network Sizes and Prediction Times per Sample}
	\resizebox{\columnwidth}{!}{
		\begin{tabular}{p{4em}P{2.5em}ccccccP{3em}}
			\toprule[1pt]
			\multirow{4}[5]{4em}{\textbf{Network}} & \multicolumn{1}{c}{\multirow{4}[5]{2.2em}{\textbf{No. of Para-meters}}} & \multicolumn{6}{P{17em}}{\textbf{Prediction Time (ms) per Sample}} & \multirow{4}[5]{1cm}{\textbf{Weight file size}} \\
			
			\multicolumn{1}{p{1cm}}{} &       & \multicolumn{6}{P{17em}}{\textbf{for Different Sample Lengths}} & \multicolumn{1}{c}{} \\
			
			\cmidrule{3-8}    \multicolumn{1}{p{1cm}}{} &       & \multicolumn{2}{P{5em}}{\textbf{0.5s}} & \multicolumn{2}{P{5em}}{\textbf{1.0s}} & \multicolumn{2}{P{5em}}{\textbf{2.0s}} & \multicolumn{1}{c}{} \\
			
			\cmidrule(r{.25em}l){3-4} \cmidrule(r{.25em}l){5-6} \cmidrule(r{.25em}l){7-8}  \multicolumn{1}{p{1cm}}{} &       & \multicolumn{1}{p{1em}}{\textbf{CPU}} & \multicolumn{1}{p{1em}}{\textbf{GPU}} & \multicolumn{1}{p{1em}}{\textbf{CPU}} & \multicolumn{1}{p{1em}}{\textbf{GPU}} & \multicolumn{1}{p{1em}}{\textbf{CPU}} & \multicolumn{1}{p{1em}}{\textbf{GPU}} & \multicolumn{1}{c}{} \\
			
			\midrule[1pt]
			\textbf{GMM} & 92,416 & 1.8   & - & 3.6   & - & 7.2   & - & 370KB \\
			
			\midrule
			\textbf{SNN} & 179,203 & 0.71  & 0.72  & 0.92  & 0.72  & 1.52  & 0.82  & 717KB \\
			
			\midrule
			\textbf{SwishNet-slim} & 5,483 & 0.77  & 1.42  & 0.88  & 1.43  & 1.12  & 1.45  & 22KB \\
			
			\midrule
			\textbf{SwishNet-wide} & 18,267 & 0.9   & 1.45  & 1.1   & 1.45  & 1.65  & 1.45  & 66KB \\
			
			\midrule
			\textbf{MobileNet} & 217,235 & 4.1   & 3.51  & 5.27  & 3.52  & 8.57  & 3.55  & 870KB \\
			\bottomrule
		\end{tabular}
	}%
	\label{tab:speed}%
\end{table}%

	The SwishNet models are much faster on the CPU than the conventional GMM approach for all clip lengths. Of all the models under consideration, SwishNet has the lowest number of parameters and the smallest weight file sizes. SwishNet-slim and SwishNet-wide are about 2.5\% and 7.5\% of the size of MobileNet respectively. Also, the memory consumed by intermediate calculations is very low due to the 1D (instead of 2D) feature maps in SwishNet. In the intermediate layers of SwishNet as few as 8/16 1D feature maps pass from one layer to the next which is even smaller in number than the original MFCC feature space (20 channels). This computational efficiency on CPU, combined with the small network size, makes SwishNet suitable for implementation on CPU-centric systems.
	\begin{table}[t]
	\centering
	\caption{Overall and Speech/Non-Speech (SNS) Classification Accuracy for Clips of Different Lengths}
	\resizebox{\columnwidth}{!}{
		\begin{tabular}{p{3em}p{4.1em}P{2.2em}P{2.2em}P{2.2em}P{2.2em}P{2.2em}P{2.2em}}
			\toprule
			\multicolumn{2}{p{8em}}{\textbf{Clip Length}} & \multicolumn{2}{M{6em}}{\textbf{0.5s}} & \multicolumn{2}{M{6em}}{\textbf{1.0s}} & \multicolumn{2}{M{6em}}{\textbf{2.0s}} \\
			
			\cmidrule(r{.25em}){1-2} \cmidrule(r{.25em}l){3-4} \cmidrule(r{.25em}l){5-6} \cmidrule(r{.25em}l){7-8} 
			\textbf{Network} &  & \textbf{Overall} & \textbf{SNS} & \textbf{Overall} & \multicolumn{1}{p{1.1em}}{\textbf{SNS}} & \textbf{Overall} & \textbf{SNS} \\
			
			\midrule[1pt]
			\textbf{GMM} &  & 96.53\% & 98.58\% & 97.33\% & 99.05\% & 97.79\% & 99.33\% \\
			
			\midrule
			\textbf{SNN} &  & 97.07\% & 98.87\% & 97.41\% & 99.13\% & 97.71\% & 99.36\% \\
			
			\midrule
			\multirow{2}{4em}{\textbf{Swish-Net-slim}} & \textbf{Undistilled} & 97.64\% & 99.24\% & 98.20\% & 99.60\% & 98.41\% & 99.76\% \\
			
			\cmidrule(l{0.5em}){2-8}          & \textbf{Distilled} & 97.52\% & 99.19\% & 98.22\% & 99.51\% & 98.57\% & 99.70\% \\
			
			\midrule
			\multirow{2}{4em}{\textbf{Swish-Net-wide}} & \textbf{Undistilled} & 97.97\% & 99.37\% & 98.32\% & 99.67\% & 98.65\% & 99.75\% \\
			
			\cmidrule(l{0.5em}){2-8}          & \textbf{Distilled} & 98.05\% & 99.45\% & 98.54\% & 99.71\% & 98.92\% & 99.84\% \\
			
			\midrule
			\multirow{2}{4em}{\textbf{Mobile-Net}} & \textbf{Random} & 98.13\% & 99.43\% & 98.53\% & 99.71\% & 98.95\% & 99.88\% \\
			
			\cmidrule(l{0.5em}){2-8}          & \textbf{Pretrained} & 98.94\% & 99.73\% & 99.24\% & 99.89\% & 99.38\% & 99.96\% \\
			\bottomrule
		\end{tabular}
	}%
	\label{tab:classperm}%
\end{table}%
	\begin{table}[t]
	\centering
	\caption{Class-wise Average F\textsubscript{1} Score for Clips of Different Lengths}
	\begin{tabular}{p{7em}p{5em}P{3em}P{3em}P{3em}}
		\toprule
		
		
		\textbf{Network} &  & \textbf{0.5s} & \textbf{1.0s} & \textbf{2.0s} \\
		
		\midrule[1pt]
		\textbf{GMM} &  & 90.63\% & 92.30\% & 93.03\%  \\
		
		\midrule
		\textbf{SNN} &  & 91.03\% & 91.61\% & 92.27\%  \\
		
		\midrule
		\multirow{2}{7em}{\textbf{SwishNet-slim}} & \textbf{Undistilled} & 93.00\% & 94.28\% & 94.84\%  \\
		
		\cmidrule(l{0.5em}){2-5}          & \textbf{Distilled} & 92.67\% & 94.50\% & 95.56\% \\
		
		\midrule
		\multirow{2}{8em}{\textbf{SwishNet-wide}} & \textbf{Undistilled} & 93.95\% & 94.61\% & 95.68\%  \\
		
		\cmidrule(l{0.5em}){2-5}          & \textbf{Distilled} & 94.05\% & 95.43\% & 96.51\%\\
		
		\midrule
		\multirow{2}{7em}{\textbf{Mobile-Net}} & \textbf{Random} & 94.22\% & 95.14\% & 96.48\% \\
		
		\cmidrule(l{0.5em}){2-5}          & \textbf{Pretrained} & 96.70\% & 97.49\% & 97.91\% \\
		\bottomrule
	\end{tabular}
	\label{tab:classf1}%
\end{table}%

\begin{table}
	\centering
	\caption{Normalized Confusion Matrices for Clips of Different Lengths (Rows: True Labels, Columns: Predicted labels, Ordering: Noise, Music, and Speech)}
	\begin{tabular}{M{5.4em}P{6.5em}P{6.5em}P{6.5em}} 
		
		\toprule
		\textbf{Network} & \textbf{0.5s} & \textbf{1.0s} & \textbf{2.0s} \\
		
		\midrule[1pt] 
		\textbf{GMM} 
		& $\begin{bmatrix}
		.79 & .19 & .02\\
		.02 & .96 & .01\\
		.01 & .01 & .98
		\end{bmatrix} $
		& $\begin{bmatrix}
		.80 & .18 & .01\\
		.02 & .97 & .01\\
		.01 & .00 & .99
		\end{bmatrix}$
		& $\begin{bmatrix}
		.79 & .18 & .02\\
		.01 & .98 & .01\\
		.00 & .00 & .99
		\end{bmatrix}$ \\
		
		\midrule
		\textbf{SNN} 
		& $\begin{bmatrix}
		.67 & .28 & .05\\
		.01 & .98 & .02\\
		.00 & .00 & 1.0
		\end{bmatrix} $
		& $\begin{bmatrix}
		.67 & .28 & .05\\
		.00 & .98 & .01\\
		.00 & .00 & 1.0
		\end{bmatrix}$
		& $\begin{bmatrix}
		.68 & .28 & .04\\
		.00 & .99 & .01\\
		.00 & .00 & 1.0
		\end{bmatrix}$ \\
		
		\midrule
		\textbf{SwishNet-slim (Undistilled)}
		& $\begin{bmatrix}
		.78 & .19 & .03\\
		.01 & .98 & .01\\
		.00 & .00 & .99
		\end{bmatrix}$ 
		& $\begin{bmatrix}
		.83 & .14 & .03\\
		.01 & .99 & .00\\
		.00 & .00 & 1.0
		\end{bmatrix}$
		& $\begin{bmatrix}
		.84 & .15 & .01\\
		.01 & .99 & .00\\
		.00 & .00 & .99
		\end{bmatrix}$\\
		
		\midrule
		\textbf{SwishNet-wide (Distilled)} 
		& $\begin{bmatrix}
		.81 & .17 & .02\\
		.01 & .99 & .00\\
		.00 & .00 & 1.0
		\end{bmatrix}$ & 
		$\begin{bmatrix}
		.86 & .13 & .01\\
		.01 & .99 & .00\\
		.00 & .00 & 1.0
		\end{bmatrix}$ 
		& $\begin{bmatrix}
		.88 & .11 & .01\\
		.01 & .99 & .00\\
		.00 & .00 & 1.0
		\end{bmatrix}$ \\
		
		\midrule 
		\textbf{MobileNet} 
		& $\begin{bmatrix}
		.90 & .09 & .01\\
		.01 & .99 & .01\\
		.00 & .00 & 1.0
		\end{bmatrix}$ 
		& $\begin{bmatrix}
		.91 & .09 & .01\\
		.00 & .99 & .00\\
		.00 & .00 & 1.0
		\end{bmatrix}$ 
		& $\begin{bmatrix}
		.94 & .06 & .00\\
		.01 & .99 & .00\\
		.00 & .00 & 1.0
		\end{bmatrix}$ \\
		\bottomrule
	\end{tabular}
	\label{tab:conmat}%
\end{table}

	The classification results are presented in Table \ref{tab:classperm}, \ref{tab:classf1} and \ref{tab:conmat}. We see that SwishNet performs better than the conventional GMM approach and also the fully connected SNN for all clip lengths. The performance of SwishNet is only overshadowed by MobileNet which is a much bigger network, pretrained on a huge dataset i.e. ImageNet. From the confusion matrices, we can see that recall for noise is low for all models, especially for smaller clip lengths, even though the speech/non-non-speech discrimination accuracy is high for all clip lengths. This result indicates that for smaller contextual windows, it is hard to differentiate noise from music. This is because, some of the noises, such as bells tolling or phones ringing are slightly musical in nature and short segments of music may sound like noise. So, a longer context is necessary for discriminating noise from music. Also, we see that although SNN can reasonably distinguish speech from non-speech, its performance in discriminating between noise and music is not at all satisfactory. This is because of the inability of an FNN to properly utilize contextual information, which is essential for distinguishing noise from music. This result justifies our use of a deep convolutional architecture which ensures proper utilization of contextual information.

	We see a significant improvement in classification performance of MobileNet when it is initialized with ImageNet weights rather than random weights. This result validates our assumption of transfer learning from ImageNet. When knowledge is distilled from MobileNet to SwishNet, we see a more noticeable improvement of performance for the wider version. For shorter clip lengths SwishNet-slim cannot benefit much from distillation because of its low capacity. The results in the following sections are given for undistilled SwishNet-slim and distilled SwishNet-wide.

	The performance levels of different models are compared in Fig. \ref{fig:perfchrat}, which demonstrates that the SwishNet models give a very good balance among size, speed and accuracy.
	\subsection{Segmentation Results}
\begin{table}[t]
	\centering
	\caption{Overall and Speech/Non-Speech (SNS) Frame-wise Segmentation Accuracy for Contextual Windows of Different Lengths (with Median Filtering)}
	\resizebox{\columnwidth}{!}{
	\begin{tabular}{p{4.5em}P{2.5em}P{2.5em}P{3em}P{2.5em}P{2.5em}P{2.5em}}
		\toprule
		\textbf{Win. Len.} & \multicolumn{2}{P{5.2em}}{\textbf{0.5s}} & \multicolumn{2}{P{5.2em}}{\textbf{1.0s}} & \multicolumn{2}{P{5.2em}}{\textbf{2.0s}} \\
		\cmidrule(r{.25em}){1-1} \cmidrule(r{.25em}l){2-3} \cmidrule(r{.25em}l){4-5} \cmidrule(r{.25em}l){6-7} 
		\textbf{Network} & \textbf{Overall} & \textbf{SNS} & \textbf{Overall} & \textbf{SNS} & \textbf{Overall} & \textbf{SNS} \\

		\midrule[1pt]
		\textbf{GMM} 			& 91.19\% & 97.33\% & 91.19\% & 97.33\% & 91.19\% & 97.33\% \\
		\midrule
		\textbf{SNN} 			& 90.38\% & 97.28\% & 90.38\% & 97.28\% & 90.38\% & 97.28\% \\
		\midrule
		\textbf{SwishNet-slim} 	& 92.70\% & 98.42\% & 93.26\% & 98.37\% & 93.08\% & 97.76\% \\
		\midrule
		\textbf{SwishNet-wide} 	& 94.09\% & 98.57\% & 94.11\% & 98.46\% & 93.55\% & 97.58\% \\
		\midrule
		\textbf{MobileNet} 		& 96.73\% & 99.37\% & 96.26\% & 98.98\% & 95.03\% & 98.31\% \\
		\bottomrule
	\end{tabular}%
    }
	\label{tab:segres}%
\end{table}%

\begin{table}
	\centering
	\caption{Normalized Confusion Matrices for Contextual Windows of Different Lengths (Rows: True Labels, Columns: Predicted labels, Ordering: Noise, Music, and Speech)}
	\begin{tabular}{p{4em}P{7em}P{7em}P{7em}} 
		
		\toprule
		\textbf{Network} & \textbf{0.5s} & \textbf{1.0s} & \textbf{2.0s} \\
		
		\midrule[1pt] 
		\textbf{GMM} 
		& $\begin{bmatrix}
		.75 & .20 & .05\\
		.04 & .95 & .02\\
		.01 & .01 & .98
		\end{bmatrix} $
		& $\begin{bmatrix}
		.75 & .20 & .05\\
		.04 & .95 & .02\\
		.01 & .01 & .98
		\end{bmatrix} $
		& $\begin{bmatrix}
		.75 & .20 & .05\\
		.04 & .95 & .02\\
		.01 & .01 & .98
		\end{bmatrix} $ \\
		
		\midrule
		\textbf{SNN} 
		& $\begin{bmatrix}
		.63 & .28 & .09\\
		.00 & .98 & .01\\
		.00 & .00 & 1.0
		\end{bmatrix} $
		& $\begin{bmatrix}
		.63 & .28 & .09\\
		.00 & .98 & .01\\
		.00 & .00 & 1.0
		\end{bmatrix} $
		& $\begin{bmatrix}
		.63 & .28 & .09\\
		.00 & .98 & .01\\
		.00 & .00 & 1.0
		\end{bmatrix} $ \\
		
		\midrule
		\textbf{SwishNet-slim}
		& $\begin{bmatrix}
		.73 & .22 & .05\\
		.01 & .99 & .01\\
		.00 & .00 & 1.0
		\end{bmatrix}$ 
		& $\begin{bmatrix}
		.75 & .20 & .05\\
		.01 & .99 & .01\\
		.00 & .00 & 1.0
		\end{bmatrix}$
		& $\begin{bmatrix}
		.76 & .18 & .06\\
		.01 & .98 & .01\\
		.00 & .01 & .99
		\end{bmatrix}$\\
		
		\midrule
		\textbf{SwishNet-wide} 
		& $\begin{bmatrix}
		.78 & .18 & .04\\
		.00 & .99 & .01\\
		.00 & .00 & 1.0
		\end{bmatrix}$ & 
		$\begin{bmatrix}
		.79 & .17 & .04\\
		.01 & .98 & .00\\
		.00 & .00 & 1.0
		\end{bmatrix}$ 
		& $\begin{bmatrix}
		.77 & .16 & .07\\
		.01 & .98 & .01\\
		.00 & .01 & .99
		\end{bmatrix}$ \\
		
		\midrule 
		\textbf{MobileNet} 
		& $\begin{bmatrix}
		.88 & .10 & .02\\
		.00 & 1.0 & .00\\
		.00 & .00 & 1.0
		\end{bmatrix}$ 
		& $\begin{bmatrix}
		.86 & .11 & .03\\
		.00 & .99 & .00\\
		.00 & .00 & 1.0
		\end{bmatrix}$ 
		& $\begin{bmatrix}
		.82 & .13 & .05\\
		.01 & .99 & .01\\
		.00 & .01 & .99
		\end{bmatrix}$ \\
		\bottomrule
	\end{tabular}
	\label{tab:segconmat}%
\end{table}
	
	The segmentation results are presented in Table \ref{tab:segres} and \ref{tab:segconmat}. Table \ref{tab:segres} shows that a 1s long contextual window with median filtering achieves the best accuracy. This is because networks with longer contextual windows are prone to making wrong predictions during transitions from one category to another. Relative performance levels of different models are similar to that for classification, with SwishNet models performing better than GMM and SNN. Table \ref{tab:segconmat} shows that noise-recall has further deteriorated for segmentation. This is because some noise periods are very short ($\sim$0.1s) in duration which are hard to detect.
	
	\subsection{Classification Results on GTZAN}

\begin{table}[!t]
	\centering
	\caption{Classification Results For 2s Long Clips Derived from GTZAN}
	\begin{tabular}{p{7em}P{4.1em}P{2.7em}P{2.7em}P{2.7em}P{2.5em}}
		\toprule
		\multirow{2}[2]{*}{\textbf{Network}} & \textbf{Overall Accuracy} & \textbf{Speech Recall} & \textbf{Music Recall} & \textbf{Average Recall} & \textbf{F\textsubscript{1} Score} \\
		
		\midrule
		\textbf{SwishNet-slim} & 98.00\% & 98.17\% & 97.84\% & 98.01\%  & 97.90\%\\
		\midrule
		\textbf{SwishNet-wide} & 98.26\% & 98.33\% & 98.20\% & 98.27\%  & 98.17\%\\
		\midrule
		\textbf{MobileNet} & 98.83\% & 98.28\% & 99.45\% & 98.86\%  & 98.77\%\\
		\bottomrule
	\end{tabular}%
	\label{tab:gtzanres}%
\end{table}%
	The music/speech classification results for 2s long clips derived from GTZAN corpus are presented in Table \ref{tab:gtzanres}. From the results, we see that even when fine-tuned on only 25\% of the new dataset, SwishNet achieves really good results. This is impressive since the GTZAN speech data have severe background noise, but the MUSAN corpus, on which the networks were trained, had almost no background noise. The recall for music is excellent even though the test data had genres of music that the networks were not even trained on. This indicates the robustness of the networks to input data.
	
	\subsection{Discussions and Trade-offs}
	SwishNet-slim is an extremely fast and lightweight network. It can be easily integrated to the front end of a real-time processing pipeline while introducing very little delay. SwishNet-wide compromises speed and size slightly in favour of higher performance and robustness which increases with distillation. The graceful performance improvement with network size also proves the scalability of our architecture and provides a very balanced trade-off between speed and performance.

	SwishNet performs much better than frame-wise approaches such as GMM an SNN because of its ability to pick up and retain information over long contextual windows. However, its performance is quite lower compared to pretrained MobileNet. Without pretraining, MobileNet is actually similar in performance to SwishNet-wide. However, when aided by transfer learning from ImageNet, it achieves excellent accuracy scores and robustness. However, while comparing our network to pretrained MobileNet, we should keep in mind that, MobileNet is a much more heavyweight and slower network compared to ours. Also, it enjoys the advantage of being pretrained on a massive dataset i.e. ImageNet. While it may be practical to use MobileNet in some off-line indexing tasks, the increased performance may not be justifiable when speed and efficiency is the priority and also when computational resources are limited. SwishNet would be a much better choice in those scenarios.

	However, it is interesting to notice that, we can distill some of the knowledge from MobileNet (initialized with ImageNet weights) to SwishNet-wide. It can be assumed that we can indirectly transfer knowledge from ImageNet which is a database of natural images to SwishNet, a model designed to classify audio, which is an interesting finding.

	\section{Conclusion}
	In this work, we proposed a novel 1D convolutional neural network -– SwishNet designed for efficient classification and segmentation of audio signals into three major categories – Speech, Music and Noise. We presented the results for two different versions of the network tuned for speed and performance respectively and compared them with conventional models such as GMM as well as a well-established 2D CNN -- MobileNet, in terms of speed and accuracy. We further showed how the performance of our network can be improved by distillation. We demonstrated that our network achieves high performance while also being less demanding on the system and introducing very little processing delay.

	
	%
	
	%
	%
	%
	%

	\ifCLASSOPTIONcaptionsoff
	\newpage
	\fi

	
	
	%
	
	\bibliographystyle{IEEEtran}
	\bibliography{mybig}

\begin{thebibliography}{10}
\providecommand{\url}[1]{#1}
\csname url@samestyle\endcsname
\providecommand{\newblock}{\relax}
\providecommand{\bibinfo}[2]{#2}
\providecommand{\BIBentrySTDinterwordspacing}{\spaceskip=0pt\relax}
\providecommand{\BIBentryALTinterwordstretchfactor}{4}
\providecommand{\BIBentryALTinterwordspacing}{\spaceskip=\fontdimen2\font plus
\BIBentryALTinterwordstretchfactor\fontdimen3\font minus
  \fontdimen4\font\relax}
\providecommand{\BIBforeignlanguage}[2]{{%
\expandafter\ifx\csname l@#1\endcsname\relax
\typeout{** WARNING: IEEEtran.bst: No hyphenation pattern has been}%
\typeout{** loaded for the language `#1'. Using the pattern for}%
\typeout{** the default language instead.}%
\else
\language=\csname l@#1\endcsname
\fi
#2}}
\providecommand{\BIBdecl}{\relax}
\BIBdecl

\bibitem{Wolfe2002}
J.~Wolfe, ``{Speech and music, acoustics and coding, and what music might be
  ‘for'},'' in \emph{Proc. 7th International Conference on Music Perception
  and Cognition}, 2002, pp. 10--13.

\bibitem{Kola2011}
J.~Kola, C.~Espy-Wilson, and T.~Pruthi, ``{Voice activity detection},''
  \emph{Merit Bien}, pp. 1--6, 2011.

\bibitem{LeCun2015}
Y.~LeCun, Y.~Bengio, and G.~Hinton, ``{Deep learning},'' \emph{nature}, vol.
  521, no. 7553, p. 436, 2015.

\bibitem{Bengio2009}
Y.~Bengio, ``{Learning deep architectures for AI},'' \emph{Foundations and
  trends{\textregistered} in Machine Learning}, vol.~2, no.~1, pp. 1--127,
  2009.

\bibitem{LeCun1995}
Y.~LeCun and Y.~Bengio, ``{Convolutional networks for images, speech, and time
  series},'' \emph{The handbook of brain theory and neural networks}, vol.
  3361, no.~10, p. 1995, 1995.

\bibitem{Krizhevsky2012}
A.~Krizhevsky, I.~Sutskever, and G.~E. Hinton, ``{Imagenet classification with
  deep convolutional neural networks},'' in \emph{Advances in neural
  information processing systems}, 2012, pp. 1097--1105.

\bibitem{VanDenOord2016b}
A.~{Van Den Oord}, S.~Dieleman, H.~Zen, K.~Simonyan, O.~Vinyals, A.~Graves,
  N.~Kalchbrenner, A.~Senior, and K.~Kavukcuoglu, ``{Wavenet: A generative
  model for raw audio},'' \emph{arXiv preprint arXiv:1609.03499}, 2016.

\bibitem{Logan2000}
B.~Logan, ``{Mel Frequency Cepstral Coefficients for Music Modeling},'' in
  \emph{ISMIR}, vol. 270, 2000, pp. 1--11.

\bibitem{Szegedy2015}
C.~Szegedy, W.~Liu, Y.~Jia, P.~Sermanet, S.~Reed, D.~Anguelov, D.~Erhan,
  V.~Vanhoucke, and A.~Rabinovich, ``{Going deeper with convolutions},'' in
  \emph{Proceedings of the IEEE conference on computer vision and pattern
  recognition}, 2015, pp. 1--9.

\bibitem{VandenOord2016a}
A.~van~den Oord, N.~Kalchbrenner, L.~Espeholt, O.~Vinyals, and A.~Graves,
  ``{Conditional image generation with pixelcnn decoders},'' in \emph{Advances
  in Neural Information Processing Systems}, 2016, pp. 4790--4798.

\bibitem{Chollet2016}
F.~Chollet, ``{Xception: Deep learning with depthwise separable
  convolutions},'' \emph{arXiv preprint}, 2016.

\bibitem{Sercu2016}
T.~Sercu, C.~Puhrsch, B.~Kingsbury, and Y.~LeCun, ``{Very deep multilingual
  convolutional neural networks for LVCSR},'' in \emph{Acoustics, Speech and
  Signal Processing (ICASSP), 2016 IEEE International Conference on}.\hskip 1em
  plus 0.5em minus 0.4em\relax IEEE, 2016, pp. 4955--4959.

\bibitem{Hershey2017}
S.~Hershey, S.~Chaudhuri, D.~P.~W. Ellis, J.~F. Gemmeke, A.~Jansen, R.~C.
  Moore, M.~Plakal, D.~Platt, R.~A. Saurous, and B.~Seybold, ``{CNN
  architectures for large-scale audio classification},'' in \emph{Acoustics,
  Speech and Signal Processing (ICASSP), 2017 IEEE International Conference
  on}.\hskip 1em plus 0.5em minus 0.4em\relax IEEE, 2017, pp. 131--135.

\bibitem{Takahashi2016}
N.~Takahashi, M.~Gygli, B.~Pfister, and L.~{Van Gool}, ``{Deep convolutional
  neural networks and data augmentation for acoustic event detection},''
  \emph{arXiv preprint arXiv:1604.07160}, 2016.

\bibitem{Deng2009}
J.~Deng, W.~Dong, R.~Socher, L.-J. Li, K.~Li, and L.~Fei-Fei, ``{Imagenet: A
  large-scale hierarchical image database},'' in \emph{Computer Vision and
  Pattern Recognition, 2009. CVPR 2009. IEEE Conference on}.\hskip 1em plus
  0.5em minus 0.4em\relax IEEE, 2009, pp. 248--255.

\bibitem{Oquab2014}
M.~Oquab, L.~Bottou, I.~Laptev, and J.~Sivic, ``{Learning and transferring
  mid-level image representations using convolutional neural networks},'' in
  \emph{Computer Vision and Pattern Recognition (CVPR), 2014 IEEE Conference
  on}.\hskip 1em plus 0.5em minus 0.4em\relax IEEE, 2014, pp. 1717--1724.

\bibitem{Simonyan2014}
K.~Simonyan and A.~Zisserman, ``{Very deep convolutional networks for
  large-scale image recognition},'' \emph{arXiv preprint arXiv:1409.1556},
  2014.

\bibitem{Howard2017}
A.~G. Howard, M.~Zhu, B.~Chen, D.~Kalenichenko, W.~Wang, T.~Weyand,
  M.~Andreetto, and H.~Adam, ``{Mobilenets: Efficient convolutional neural
  networks for mobile vision applications},'' \emph{arXiv preprint
  arXiv:1704.04861}, 2017.

\bibitem{Zoph2017}
B.~Zoph, V.~Vasudevan, J.~Shlens, and Q.~V. Le, ``{Learning transferable
  architectures for scalable image recognition},'' \emph{arXiv preprint
  arXiv:1707.07012}, 2017.

\bibitem{Hinton2015}
G.~Hinton, O.~Vinyals, and J.~Dean, ``{Distilling the knowledge in a neural
  network},'' \emph{arXiv preprint arXiv:1503.02531}, 2015.

\bibitem{Snyder2015}
D.~Snyder, G.~Chen, and D.~Povey, ``{Musan: A music, speech, and noise
  corpus},'' \emph{arXiv preprint arXiv:1510.08484}, 2015.

\bibitem{musanpage}
\BIBentryALTinterwordspacing
``{MUSAN - OpenSLR [Online]}.'' [Online]. Available:
  \url{http://www.openslr.org/17/}
\BIBentrySTDinterwordspacing

\bibitem{Tzanetakis2002}
G.~Tzanetakis and P.~Cook, ``{Musical genre classification of audio signals},''
  \emph{IEEE Transactions on speech and audio processing}, vol.~10, no.~5, pp.
  293--302, 2002.

\bibitem{Ephraim1985}
Y.~Ephraim and D.~Malah, ``{Speech enhancement using a minimum mean-square
  error log-spectral amplitude estimator},'' \emph{IEEE transactions on
  acoustics, speech, and signal processing}, vol.~33, no.~2, pp. 443--445,
  1985.

\bibitem{Klambauer2017}
G.~Klambauer, T.~Unterthiner, A.~Mayr, and S.~Hochreiter, ``{Self-normalizing
  neural networks},'' in \emph{Advances in Neural Information Processing
  Systems}, 2017, pp. 971--980.

\bibitem{Kingma2014}
D.~P. Kingma and J.~Ba, ``{Adam: A method for stochastic optimization},''
  \emph{arXiv preprint arXiv:1412.6980}, 2014.

\bibitem{Loshchilov2016}
I.~Loshchilov and F.~Hutter, ``{SGDR: stochastic gradient descent with
  restarts},'' \emph{arXiv preprint arXiv:1608.03983}, 2016.

\bibitem{Abadi2016}
M.~Abadi, P.~Barham, J.~Chen, Z.~Chen, A.~Davis, J.~Dean, M.~Devin,
  S.~Ghemawat, G.~Irving, and M.~Isard, ``{Tensorflow: a system for large-scale
  machine learning},'' in \emph{OSDI}, vol.~16, 2016, pp. 265--283.

\end{thebibliography}
	
	%
	%
	
	%
	
%
%
%
	
	
	

\end{document}